\documentclass[preprint,3p,11pt]{elsarticle}
\usepackage{lineno,hyperref}
\usepackage{times}
\usepackage{soul}
\usepackage{url}
\usepackage[utf8]{inputenc}
\usepackage[small]{caption}
\usepackage{graphicx}
\usepackage{amsmath}
\usepackage{amsthm}
\usepackage{booktabs}
\usepackage{algorithm}
\usepackage{algorithmicx, algpseudocode}
\usepackage{dsfont}
\usepackage{amsmath}
\usepackage{nicefrac}
\usepackage{caption}
\DeclareMathOperator*{\argmax}{arg\,max}

\modulolinenumbers[5]

\journal{Artificial Intelligence}

\begin{document}

\begin{frontmatter}

\title{
The First AI4TSP Competition: 
Learning to Solve Stochastic 
Routing
Problems}



\author[TUe]{Laurens Bliek}
\author[TUe]{Paulo da Costa}
\author[TUe]{Reza Refaei Afshar}
\author[TUe]{Yingqian Zhang}
\author[TUD]{Tom Catshoek}
\author[TUD]{Dani\"el Vos}
\author[TUD]{Sicco Verwer}

\address[TUe]{Eindhoven University of Technology, Netherlands}
\address[TUD]{Delft University of Technology, Netherlands}


\author[McG]{Fynn Schmitt-Ulms}
\author[BI]{Andr\'{e} Hottung}
\author[GE]{Tapan Shah}
\author[SH]{Meinolf Sellmann}
\author[BI]{Kevin Tierney}

\address[McG]{McGill University, Canada}
\address[BI]{Bielefeld University, Germany}
\address[GE]{General Electric, USA}
\address[SH]{Shopify, Canada}

\author[UdeM]{Carl Perreault-Lafleur}
\author[UdeM]{Caroline Leboeuf}
\author[UdeM]{Federico Bobbio}
\author[UdeM]{Justine Pepin}
\author[UdeM]{Warley Almeida Silva}
\address[UdeM]{Université de Montréal, Canada}

\author[RG]{Ricardo Gama}
\address[RG]{Polytechnic Institute of Viseu, Portugal}
\author[HF]{Hugo L. Fernandes}
\address[HF]{Rockets of Awesome, New York City, USA}

\author[THK]{Martin Zaefferer}
\author[UM]{Manuel L{\'{o}}pez{-}Ib{\'{a}}{\~{n}}ez}
\author[TP]{Ekhine Irurozki}

\address[UM]{University of M{\'{a}}laga, Spain}
\address[TP]{Telecom Paris, France}
\address[THK]{TH K\"oln, Germany}






\begin{abstract}
This paper reports on the first international competition on AI for the traveling salesman problem (TSP) at the International Joint Conference on Artificial Intelligence 2021 (IJCAI-21). The TSP is one of the classical combinatorial optimization problems, with many variants inspired by real-world applications. 
This first competition asked the participants to develop algorithms to solve a time-dependent orienteering problem with stochastic weights and time windows (TD-OPSWTW). It focused on two types of learning approaches: surrogate-based optimization and deep reinforcement learning. In this paper, we describe the problem, the setup of the competition, the winning methods, and give an overview of the results.
The winning methods described in this work have advanced the state-of-the-art in using AI for stochastic routing problems.
Overall, by organizing this competition we have introduced routing problems as an interesting problem setting for AI researchers.
The simulator of the problem has been made open-source and can be used by other researchers as a benchmark for new AI methods.
\end{abstract}

\begin{keyword}
AI for TSP competition \sep Travelling salesman problem \sep Routing problem \sep Stochastic combinatorial optimization \sep Surrogate-based optimization \sep Deep reinforcement learning 
\end{keyword}

\end{frontmatter}


\section{Introduction}

Many real-world optimization problems are combinatorial optimization problems (COPs) with an objective to find an optimal solution among a finite set of possible solutions. COPs are proved to be NP-Complete, thus solving them to optimality is computationally expensive and mostly impractical for large instances. COPs have been studied extensively in various research communities, including discrete mathematics, theoretical computer science, and operations research. 
An efficient way of finding acceptable solutions for COPs is through heuristic approaches. The time complexity of heuristics is mainly polynomial, although they may provide solutions that are far from optimal. Besides, these approaches must be redesigned if the problem assumption and settings are changed. 
Recent years have seen rapidly growing interest in using machine learning (ML) to 
dynamically learn heuristics and find close-to-optimal solutions for COPs \cite{bengio2021machine}. Among COPs, routing problems such as the traveling salesman problem (TSP) are well-known, and they emerge in many real-life applications. The TSP has several variants that include uncertainty, making the problem  challenging for traditional exact and heuristic algorithms. TSP and its variants are some of the most well-studied COPs in the ML literature. Previous works on deep neural network approaches for routing problems have focused on learning to construct good tours  \cite{vinyals2015pointer,joshi2019efficient,Bello2017NeuralCO, kool2018attention, ma2019combinatorial, nazari2018reinforcement,deudon2018learning,pomo,fu2020generalize, kool2021deep, delarue2020reinforcement, GAMA2021} and on learning to search for good solutions \cite{hottung2020neural, pmlr-v129-costa20a, da2021learning, wu2019learning, lu2020learning,chen2019learning, xin2021neurolkh,li2021learning, gao2020learn, sui2021learning,kim2021learning,hottung2021learning}, leveraging supervised and deep reinforcement learning (DRL). Other approaches considered surrogate-based optimization (SBO)~\cite{bliek2019black, BNAIC2020paper,Namazi2020SurrogateAO,Fang2020SurrogateAssistedGA,Bracher2021LearningSF,Ardeh2020AGW}, using ML models to guide the search for good tours.

In this competition, the participants solve a variant of TSP using ML methods. The selected variant of TSP contains stochastic weights, where the cost of traveling between two nodes might be different at different times. Each node also has a prize, and collecting the prize depends on the arrival time of an agent. These assumptions make this variant of TSP similar to real-life problems. For example, in real life, the required time to travel from one city to another depends on road construction work and traffic jams. Moreover, visiting a location is usually assigned with time bounds that must be respected. To solve this problem variant, the participants must use one of two ML methods: SBO or DRL. Both of these methods have shown considerable promise generating solutions for routing problems in previous works. 

We emphasize that the primary goal of this competition is to bring new surrogate-based and DRL-based approaches into practice for solving a difficult variant of TSP. This is done by attracting ML researchers and challenging them to solve this difficult routing problem. The solutions may be built upon existing work adapted for the particular TSP variant. Although some previous work has focused on prize collecting (orienteering) problems or stochastic weights, few researchers take the combination of these assumptions into account. This motivates us to establish a platform that provides the opportunity for AI researchers to develop SBO and DRL approaches for solving a well-known routing problem. As a byproduct, the competition provides several winning methods and a simulator for generating problem instances that researchers can use to benchmark their ML-based approaches. In summary, the objective of organizing this competition is threefold: (1) to introduce routing problems as an interesting problem setting for ML researchers; (2) to advance the state-of-the-art in using ML for routing problems; and (3) to provide a challenging problem and a simulator for researchers to benchmark their ML-based approaches.  




We divide the competition in two tracks, each requiring different knowledge from sub-fields of AI:
\begin{itemize}
    \item Track 1 (SBO): 
    Given one instance, previously tried tours, and the total reward (sum of the prizes collected in a tour) for those tours, the goal is to learn a model predicting the reward for a new tour. Then an optimizer finds the tour that gives the best reward according to that model, and that tour is evaluated, giving a new data point. Then the model is updated, and this iterative procedure continues for a fixed number of steps.
    Over time, the model becomes more accurate, giving better and better tours.
    This procedure is used in SBO algorithms such as Bayesian optimization~\cite{Shahriari2016TakingTH}.
    
    \item Track 2 (DRL): We consider an environment (simulator) that can generate a set of multiple instances $\mathcal{I}$ following the same generating distribution. We expect as output (partial) solutions containing the order at which the nodes should be visited. The environment returns general instance features and the stochastic travel time for traversing the last edge in a given solution. The goal is to maximize the prizes collected while respecting time-related constraints over multiple samples of selected test instances. This procedure is related to neural combinatorial optimization~\cite{Bello2017NeuralCO}.
    
\end{itemize}

The first competition, named AI4TSP competition, was an IJCAI-21 (International Joint Conference on Artificial Intelligence) competition. It ran from May 27 till July 12, 2021, and was organized by the Delft University of Technology and the Eindhoven University of Technology. 
By the deadline of the final test phase, we had received four submissions in the SBO track and three submissions in the DRL track. The submissions are tested on up to $1,\!000$ problem instances with up to $200$ nodes and the winners are determined by ranking the total quality of their solutions. The results of the competition have been officially announced in the Data Science Meets Optimization (DSO) workshop, which was co-located with IJCAI-21. The code for the competition can be found online\footnote{  \url{https://github.com/paulorocosta/ai-for-tsp-competition}.}.

\section{Problem description and methodology}

Both tracks look at the time-dependent orienteering problem with stochastic weights and time windows (TD-OPSWTW)~\cite{VerbeeckC2016StST}. This problem is similar to the traveling salesman problem (TSP), where nodes need to be visited while respecting a maximum tour time and opening and closing times of the nodes in order to maximize some measure of rewards. We detail the problem in the section below.

\subsection{TD-OPSWTW}
In the TSP, the goal is to find the tour with the smallest cost that visits all locations (customers) in a network exactly once. However, in practical applications, one rarely knows all the travel costs between locations precisely. Moreover, there could be specific time windows at which customers need to be served, and certain customers can be more valuable than others. Lastly, the salesman is often constrained by a maximum capacity or travel time, representing a limiting factor in the number of nodes that can be visited.

In this competition, we consider a more realistic version of the classical TSP, i.e., the TD-OPSWTW \cite{VerbeeckC2016StST}.
In this formulation, the stochastic travel times between locations are only revealed as the salesman travels in the network. The salesman starts from a depot and must return to the depot at the end of the tour. Moreover, each node (customer) in the network is assigned a prize, representing how important it is to visit a given customer on a given tour. Each node has associated time windows. We consider that a salesman may arrive earlier at a node without compromising its prize, but the salesman must wait until the opening time to serve the customer. Lastly, the tour must not violate a total travel time budget while collecting prizes in the network. The goal is to collect the most prizes in the network while respecting the time windows and the total travel time of a tour allowed to the salesman.

\paragraph{Node locations} More formally, a TD-OPSWTW problem instance is defined as a set of $n$ nodes that make a complete graph. The $x$ and $y$ coordinates of the nodes in a 2D space are randomly generated integers between two limits. For a node $i$, the limits are input parameters having the default values of $l_x = \{0, 200\}$ and $l_y = \{0, 50\}$ for $x_i$ and $y_i$, respectively. 

\paragraph{Travel times} The noisy travel times $t_{i,j} \in \mathds{R}, \, \forall i, j \in \{1, \ldots, n\}$ between node $i$ and $j$ are obtained by first computing their Euclidean distance $d_{i,j}$ rounded to the closest integer. Later, this distance is multiplied by a noise term $\eta$ following a discrete uniform distribution $\mathcal{U}\{1,100\}$ normalized by a scaling factor $\beta = 100$, i.e., $\tau_{i,j} = d_{i,j}\frac{\eta }{\beta} $, where $t_{i,j}$ are samples from $\tau_{i,j}$. Hence, the travel time between $i$ and $j$ may be different in different samples. 

\paragraph{Time windows}
Each node $i$ has a time window denoted by its lower bound $\{l_i \in \mathds{N}\}_{i=1}^n$ and upper bound $\{h_i \in \mathds{N}\}_{i=1}^n$. The time windows are generated around the times to begin service at each node of a second nearest neighbor TSP tour. In more detail, let $d^{\text{2nn}}_i$ be the time of visiting the $i$-th node in the second nearest neighbor tour, assuming maximum travel times between nodes. The left side of the time window is a randomly generated number between $d^{\text{2nn}}_i-w$ and $d^{\text{2nn}}_i$ where $w$ is an arbitrary integer. Similarly, the right side of the time window is a random number between $d^{\text{2nn}}_i$ and $d^{\text{2nn}}_i+w$. In our problem setting, $w \in \{20, 40, 60, 80$, $100\}$ \cite{dumas1995optimal}.

\paragraph{Prizes}
Each node has an associated prize. The prize $ \{p_i \in \mathds{R}\}^n_{i=1}$ of node $i$ is describes the importance of visiting that node within its time window. The prizes are increasing with distance between node $i$ and the first node of the tour (depot). In more detail, the prize of each node is determined according to the (rounded) $L_2$ distances between the nodes and depot. That is,  $p_i = \Big(1 + \big\lfloor 99 \cdot \frac{  d_{1, i} }{\max_{j=1}^{n} d_{1,j}}  \big\rfloor \Big) /100$, where $d_{1,i}$ is the euclidean distance (maximum travel time) from the depot to node $i$. This prize structure results in challenging instances as it places nodes with higher prizes away from the depot  \cite{fischetti1998solving}.

\paragraph{Constraints}  Each problem instance has a maximum tour length $T$ that determines the maximum time allowed to be spent on a tour. For each instance, we sample the max tour length $T$ from a discrete uniform distribution $\mathcal{U}\{T_{\min}, T_{\max} \}$, where $T_{\min} = 2 \cdot (\max_{j=1}^{n} d_{1,j})$ and $T_{\max} = \max (2 \cdot T_{\min}, \lceil \frac{1}{2}d^{\text{nn}} \rceil)$, and $d^{\text{nn}}$ is the tour cost of the nearest neighbor TSP solution with maximum travel times. Note that $T_{\min}$ is defined such that it is possible to go from the depot to the farthest node and back. $T_{\max}$ is defined as the maximum between twice the $T_{\min}$ time and half the nearest neighbor TSP tour cost to ensure that instances are still challenging with feasible, but not excessively large values of $T$. 
Moreover, solutions must respect the time windows of each node, i.e., $[l_i, h_i]$. That is, if a tour arrives earlier than the opening time of that node, it must wait until the opening time to depart the node. A tour is considered infeasible if the arrival time is higher than the closing time of a node. Note that, unlike the latter case, it is still possible to collect the prizes of a node if arriving earlier than the opening time of the time window.

\paragraph{Penalties} We treat each violation of the constraints of the problem in the form of penalties $\{e_i \in \mathds{R}\}^n_{i=1}$. 
All solutions (tours) that take longer than $T$ are penalized by $e_i = -n$, incurred at the node $i$ at which the violation first occurred. Moreover, each time window violation from above incurs a penalty of $e_i = -1$ at the current node $i$ at which the violation occurred. \\


The stochastic travel times, time-dependent constraints and prizes make this problem difficult to solve with traditional solvers. In our implementation, a problem instance is a complete graph with a particular depot (node:1). The nodes are fixed in a 2D space; however, their travel times are noisy, i.e., they can vary in different runs. 
The goal of the problem is to find a tour such that the total reward is maximized. Note that prizes, penalties, node coordinates, time window bounds and the maximum allowed tour time $T$  are known and given in the instance.


To summarize, the main differences between the problem in this competition and the TSP are:
\begin{itemize}
    \item Not all nodes need to be visited -- it is allowed to never visit some nodes.
    \item Visiting a node after the node's opening time and before its closing time gives a prize.
    \item Visiting a node after its closing time gives a penalty.
    \item When visiting a node before its opening time, the agent has to wait until the node opens.
    \item The time it takes to travel from one node to the other is 
    stochastic.
    \item  The travel times do not directly appear in the objective function. 
    \item The only metric that matters is the sum of collected prizes (penalties).
\end{itemize}

\begin{figure}[tb]
    \centering
    \begin{center}
    \includegraphics[width=0.6\columnwidth]{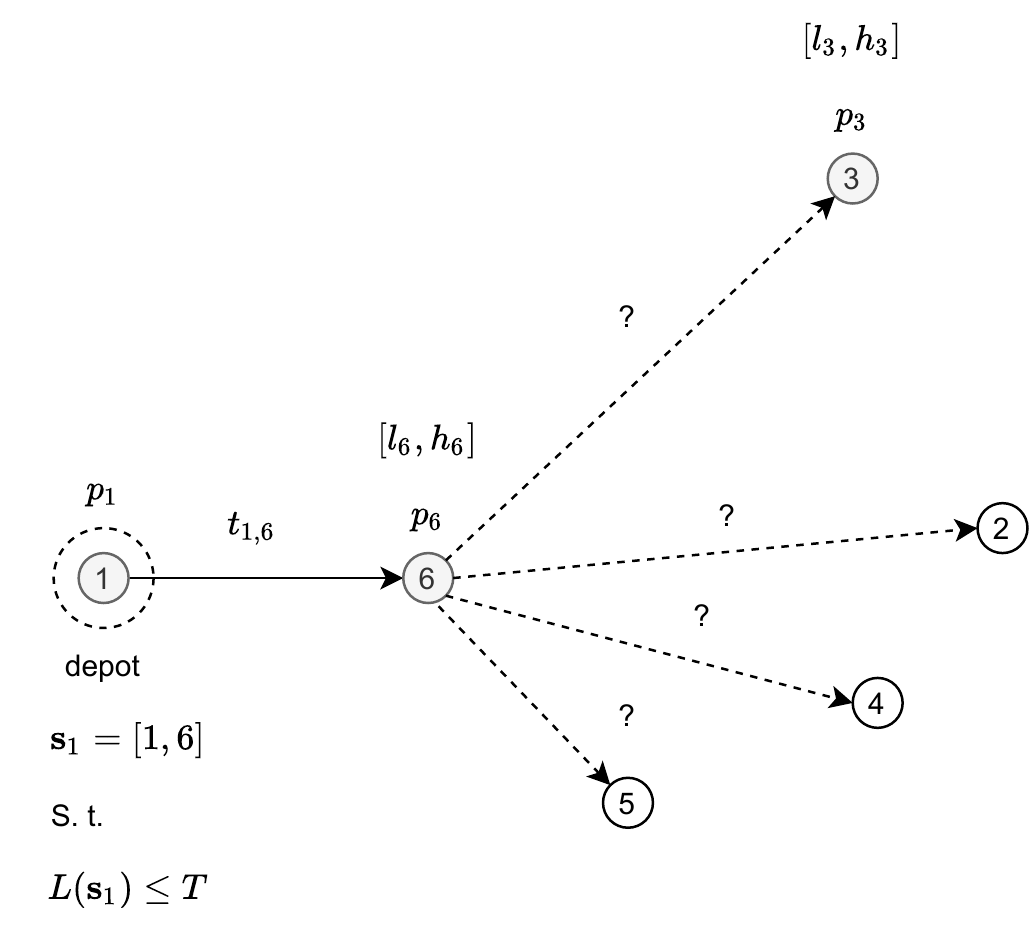}    
    \end{center}
     \caption{An instance of the TD-OPSWTW.}
      \label{fig:problem}
\end{figure}

Figure \ref{fig:problem} shows an example of a next node visitation decision that has to be made by a policy visiting $n=6$ nodes. In the figure, a tour has visited nodes 1 (depot) and 6 with travel time $t_{1,6}$, which is revealed after visiting node 6. At this current decision moment, we need to choose the next node to visit. The decision should consider the prizes of each node, the time windows, and the total remaining travel time when selecting the next node (in this case, node 3 is selected).

Moreover, when the salesman decides to arrive at a node $i$ earlier than the earliest service time $l_i$, the travel time gets shifted to the beginning of the time window. For example, if the travel time between the depot (node 1) and node 6 is lower than $l_6$, the salesman must wait until $l_6$ to depart from that node. This information becomes available as soon as the salesman arrives at node 6. Lastly, a tour must always return to the depot, and this travel time is also included in the maximum allowed tour time. 


\subsection{Track 1: surrogate-based optimization}

The goal of Track 1 is to solve an optimization problem related to one instance of the TD-OPSWTW problem, finding the tour that maximizes the total reward.
The total reward of a tour, which is the sum of all collected prizes and penalties, can be represented as a black-box function $f(\mathbf s,I)$, taking as input the instance $I$ and a tour $\mathbf s$.
The optimization problem is then denoted as:
\begin{equation}
    \mathbf s^* = \argmax_{\mathbf s} \mathds{E}[f(\mathbf s, I)] 
    \label{eq:expectedobj}
\end{equation}
for a given instance $I$.
We use the expected value because the simulator is stochastic: it can give different rewards even if the same tour is evaluated multiple times.
The expected value for a tour $\mathbf s$ is approximated by evaluating $f(\mathbf s, I)$ for that tour 10,000 times and calculating the average total reward.
This computation takes multiple seconds on standard hardware.
Therefore, the problem can be seen as an expensive optimization problem.
Surrogate-based optimization methods, such as Bayesian optimization~\cite{Shahriari2016TakingTH}, which approximate the expensive objective using online supervised learning, are known to perform well on this type of problem.

The tour $\mathbf s$ indicates the order to visit the nodes in the network.
It has to take on the specific form $\mathbf s = [1, s_1, \ldots, s_n]$, with $n$ the number of nodes and $s_1, \ldots, s_n$ containing all integers from $1$ to $n$.
This means that the number $1$ will appear twice in the solution.
As this number indicates the starting node, the tour consists of starting from the starting node, visiting any number of nodes, then returning to the starting node at some point.
Any nodes that appear in the tour after returning to the starting node are ignored.


SBO algorithms approximate the black-box function $f$ in every iteration with a surrogate model $g$ (the online learning problem), then optimize $g$ instead (the optimization problem).
Both the results of learning and optimization become better with each simulator call as more data becomes available.
The problem is typically split into two sub-problems that are solved every time a tour is given as input to the simulator:
\begin{enumerate}
    \item Given the tours tried up until now and their corresponding rewards, learn a model to predict how promising any new tour would be.
    \item Optimize the model of the previous step to suggest the most promising tour to try next. Then this tour is given as input to the simulator.
\end{enumerate}

The first step can be seen as an online learning problem, where new data comes in at every iteration, and rewards need to be predicted.
It also corresponds to the concept of an acquisition function in Bayesian optimization.
In step 2, standard optimization methods, such as gradient descent, can be used.

\subsubsection{Baseline}

As a baseline, we provide an implementation of a standard Bayesian optimization algorithm using Gaussian processes~\cite{Shahriari2016TakingTH}.
For this implementation, we use the \texttt{bayesian-optimization} Python package~\cite{bayesianoptimizationpython}
, after transforming the input space using the approach in~\cite{bliek2019black} and rounding solutions to the nearest integer.
The expectation is that such a baseline method does not perform well on the problem due to the combinatorial search space, the many constraints, and the possibility of using noisy, but low-cost approximations of the objective. The competition participants were requested to develop new methods that were more suitable for the problem than existing baseline algorithms.

\subsection{Track 2: Deep reinforcement learning}

In the DRL track, we are interested in a (stochastic) policy $\pi$ mapping states to action probabilities. A policy in the TD-OPSWTW selects the next node to be visited, given a sequence of previously visited nodes. To cope with the stochastic travel times, the policy must be adaptive. Therefore, the policy needs to consider the instance information to construct tours dynamically that respect the time windows of nodes and the total tour time allowed for the instance. Note that, unlike Track 1, we are interested in general policies applicable to any instance of the TD-OPSWTW in the training distribution. 

More formally, we adopt a standard Markov decision process (MDP) $\mathcal{M} = \langle \mathcal{S}, \mathcal{A}, \mathcal{P}, r \rangle$ where $\mathcal{S}$ is the state space, $\mathcal{A}$ is the action space, $\mathcal{P}(\mathbf{s}^\prime | \mathbf{s}, a)$ is the transition distribution after taking action $a$ at state $\mathbf{s}$, $r(\mathbf{s},a)$ is the reward function. 
Where we model a state $\mathbf{s}$ as partial and complete tours, the action space as the remaining nodes to be visited and the rewards as the sum of prizes and penalties collected at each step. Thus, the main objective is to find a policy $\pi^*$ such that 
\begin{equation}
    \pi^* = \argmax_\pi \mathds{E}_{I \sim \mathcal{P}(I)}\Big[\mathcal{L}(\pi|I)\Big], 
    \label{eq:J_pi_real}
\end{equation}
where the instance $I$ is sampled from a distribution $\mathcal{P}(I)$ and
\begin{equation}
     \mathcal{L}(\pi|I) = \mathds{E}_{\pi} \Bigg[ \sum_{i=0}^{n-1} r(\mathbf{s}_{i}, a_i) \Bigg],
    \label{eq:J_pi}
\end{equation}
where $\mathbf{s}_{i}$ is the state at decision epoch $i$, for example, a partial tour until node $s_i$, i.e., $\mathbf{s}_{i} = [s_0, s_1, \ldots, s_i]$, assuming that we always start from the depot, i.e., $s_0=1$. Note that the reward after taking action $a_i$ is given by $r(\mathbf{s}_{i}, a_i) = p_{a_i} + e_{a_i}$ if the tour has not returned to the depot, and $0$ otherwise.

\subsubsection{Baseline}

We provide a baseline to the RL track based on neural combinatorial optimization \cite{Bello2017NeuralCO}. Note that this approach is not adaptive and may not perform well in the given task as it only uses the coordinates and prizes to make decisions on complete tours. Moreover, it is not tailored for stochastic problems as it does not consider the online travel times and maximum tour budget information while traversing the tours. Participants were requested to develop new methods better suited for exploiting the complete information from the instances, including the travel times, revealed at each new location in the tour and the maximum travel budget. 



\section{Technical setup and evaluation}

\subsection{Problem instances}
\label{sec:instances}

A set of problem instances was generated and provided for the participants. Each problem instance contains $n$ nodes in 2D space. Each node has an $x$-coordinate, a $y$-coordinate, lower and upper bounds of its time window, prizes and the maximum tour time. The source code of the competition was provided to participants. Thus,  everyone had access to how instances and environments were generated, and participants could inspect each of its components. The implementation of the source code of the competition was done using the Python programming language. 

In more detail, the \texttt{InstanceGenerator} (Algorithm \ref{algo:instancegen}) class generates the features of the problem instances. This class is initialized by the number of nodes, limits of $x$ and $y$ coordinates, $w$, and a random seed. The problem instances are generated in three steps. First, the coordinates of the nodes are generated randomly between two intervals $l_x$ and $l_y$. 
Note that the location of the nodes is fixed and given as input; however, the travel times between nodes are subject to change according to noisy values.

\begin{algorithm}[ht]
\begin{algorithmic}
\Require number of nodes $n$; limits $l_x, l_y$; time window size $w$
\For{$i=1, \ldots, n$}
    \State $x_i \sim \mathcal{U}\{l_x\}$
    \State $y_i \sim \mathcal{U}\{l_y\}$
\EndFor
\State $\mathbf{x} \gets [x_1, \ldots, x_n]$,  $\mathbf{y} \gets [y_1, \ldots, y_n]$
\State $\mathbf{l}, \mathbf{h}, \mathbf{D}$ $\gets $ \texttt{TWGenerator}($\mathbf{x}$, $\mathbf{y}$, $w$)
\State $\mathbf{p}, T $ $\gets $ \texttt{PrizeGenerator}($\mathbf{x}$, $\mathbf{y}$, $\mathbf{D}$)
\State $I = [\mathbf{x}, \mathbf{y}, \mathbf{l}, \mathbf{h}, \mathbf{p}, T, \mathbf{D}]$
\State \Return $I$
 \caption{\texttt{InstanceGenerator}}
  \label{algo:instancegen}
\end{algorithmic}
\end{algorithm}


Second, the time window of each node is generated around the time of visiting that node in the second nearest neighbor tour in the \texttt{TWGenerator} class (Algorithm \ref{algo:twgen}). An instance of this class receives the node coordinates of a TD-OPSWTW instance and a value $w$, computes the $L_2$ distances between nodes, and returns the time windows for each node based on the second nearest neighbor TSP tour. 

\begin{algorithm}[ht]
\begin{algorithmic}
\Require node coordinates $\mathbf{x} = [x_1, \ldots, x_n]$, $\mathbf{y} = [y_1, \ldots, y_n]$;  time window size $w$

\For{$i=1, \ldots, n$}
\For{$j=1, \ldots, n$}
\State $d_{i,j} \gets L_2(x_i, y_i, x_j, y_j)$
\EndFor
\EndFor
\State $\mathbf{D} \gets [d_{1, 1}, \ldots, d_{n,n}]$
\State $\mathbf{s}^{\text{2nn}}, \mathbf{d}^{\text{2nn}}  \gets \texttt{GetSecondNearestNeighbor}(\mathbf{D})$ \Comment{ $2^{\text{nd}}$NN tour and its time (dist.) node-by-node}
\State $t_1 \gets 0$, $l_1 \gets 0$
\State $h_1 \gets \lceil  \max_{j=1}^n d^{\text{2nn}}_{j} + w \rceil$
\For {$j=2, \ldots, n$}
    \State $i \gets \mathbf{s}^{\text{2nn}}_j$
    \State $l_i \sim \mathcal{U}\{ \max( 0, d^{\text{2nn}}_{j}  - w  ), d^{\text{2nn}}_{j} \}$
    \State $h_i \sim \mathcal{U}\{ \max( 0, d^{\text{2nn}}_{j}), d^{\text{2nn}}_{j} + w \}$

\EndFor
\State $\mathbf{l} \gets [l_{1}, \ldots, l_{n}]$
\State $\mathbf{h} \gets [h_{1}, \ldots, h_{n}]$

\State \Return $\mathbf{l}, \mathbf{h}, \mathbf{D} $
 \caption{\texttt{TWGenerator}}
  \label{algo:twgen}
\end{algorithmic}
\end{algorithm}

Third, the prize of each node is determined according to the $L_2$ distances between the nodes and depot. The prizes are generated upon calling the \texttt{PrizeGenerator} class (Algorithm \ref{algo:prizegen}). This class takes as input the node coordinates and the $L_2$ distance between nodes, i.e., the maximum travel time, and outputs prizes based on the distance between nodes and the depot. That is, nodes farther from the depot have larger prizes. Moreover, this class generates the maximum allowed tour budget of $T$, preventing tours from obtaining a very big total prize. Participants also had access to the $L_2$ distance between nodes, which corresponds to the maximum possible travel times between nodes in our setup. 

\begin{algorithm}[ht]
\begin{algorithmic}

\Require node coordinates $\mathbf{x} = [x_1, \ldots, x_n]$, $\mathbf{y} = [y_1, \ldots, y_n]$;  maximum time matrix $\mathbf{D}$
\State $d^{\text{nn}}  \gets \texttt{GetNearestNeighborCost}(\mathbf{D})$ \Comment{$d^{\text{nn}}$ is the total travel time of the NN tour}
\State $T_{\min} = 2 \cdot (\max_{j=1}^{n} d_{1,j})$
\State $T_{\max} = \max (2 T_{\min}, \lceil \frac{1}{2}d^{\text{nn}} \rceil)$
\State $T \sim \mathcal{U}\{T_{\min}, T_{\max} \} $
\For{$i = 1, \ldots, n$}
\State $p_i = \frac{\Big(1 + \big\lfloor 99 \cdot \frac{  d_{1, i} }{\max_{j=1}^{n} d_{1,j}}  \big\rfloor \Big)}{100}$
\EndFor
\State $\mathbf{p} \gets [p_1, \ldots, p_n]$
\State \Return $\mathbf{p}, T $
 \caption{\texttt{PrizeGenerator}}
  \label{algo:prizegen}
 
\end{algorithmic}
\end{algorithm}

As a simple example, assume that there are 4 nodes in an instance, each with a time window, a prize and a maximum travel budget. An illustration of this example is shown in Table \ref{tbl:instance}. Each row of this table corresponds to a particular node, and the columns are defined as follows: \textsc{CUSTNO} ($i$) is an integer identifier for the nodes. \textsc{XCOORD} and \textsc{YCOORD} ($x_i, y_i$) are the coordinate of a node. TW\_LOW and TW\_HIGH ($l_i, h_i$) are the left sides and right sides of the time window for each node. PRIZE ($p_i$) is the prize of each node. Finally, MAX\_T ($T$) is the maximum travel budget. A possible tour for this example is $1\rightarrow{2}\rightarrow{3}\rightarrow{4}\rightarrow{1}$. The total time of this tour is 187, and the time of visiting nodes 1, 2, 3 and 4 are 0, 13, 50 and 118, respectively. At time 13, a tour visits node 2; however, it needs to wait until time 102 to collect the prize of this node. If the tour leaves node 2 after collecting its prize, it gets to node 3 after its time window. Therefore, it misses the prize of node 3 and incurs a penalty of $-1$. Then, it can get to node 4 within its time window and collect its prize. Therefore, the total collected prize for this tour is $0.19$.

\begin{table*}[ht]
    \centering
    \begin{tabular}{ccccccc}
        \toprule
        CUSTNO & XCOORD & YCOORD & TW\_LOW & TW\_HIGH & PRIZE & MAX\_T\\
        \toprule
        1 & 47 & 24 & 0 & 285 & 0.0 & 256 \\ 
        \midrule
        2 & 38 & 15 & 102 & 198 & 0.19 & 256 \\
        \midrule
        3 & 53 & 49 & 9 & 52 & 0.38 & 256 \\
        \midrule
        4 & 116 & 23 & 30 & 137 & 1.0 & 256\\
        \bottomrule

    \end{tabular}
    \caption{A sample problem instance with 4 nodes.}
    \label{tbl:instance}
\end{table*}

\subsection{Environments}
\label{sec:envs}

\subsubsection{Track 1}

The environment for Track 1 (\texttt{Env}) receives as input the instance information, containing the node coordinates, time windows, prizes, maximum allowed travel time and the maximum travel times between nodes. 
The environment implements a method \texttt{check\_solution} that takes as input a complete tour, i.e., starting and ending at the depot, returning the total reward and tour time of that tour.

\subsubsection{Track 2}

The environment for Track 2 (\texttt{EnvRL}) serves a similar purpose as the one from Track 1. Here, however, participants can interact with the environment on a node-by-node basis. That is, the method \texttt{step} expects a single node as input and builds a solution node-by-node. In doing so, the method returns the total tour time, travel time of the previous edge, rewards and penalties at each step, a feasibility indicator, and whether the tour is complete, i.e., if it has returned to the depot. This allows participants to consider the dynamics of the problem, considering the sampled travel times while constructing a solution.  

\subsection{Evaluation}


The participants were evaluated over several generated instances (see Section \ref{sec:instances}) for each track of the competition. The competition was split into two phases, validation and test, detailed below.

\paragraph{Validation phase}

In Track 1, the teams were given a single \textit{instance} containing $n=55$ nodes, i.e., a single problem containing a set of nodes and other instance information (see Section \ref{sec:instances}). During this phase, participants were allowed to test their methods on this instance without any cap on the number of evaluations. At the end of the validation phase, the performance of each team was evaluated on 10,000 Monte Carlo samples from this instance, sampling different travel times in each new sample. Performance was measured considering the sum of prizes and penalties of the proposed tour for each sample averaged over the 10,000 experiments. Participants were given a random seed to reproduce the same randomness in the experiments, such that the difference in results (if any) is from the performance of the algorithms.

For Track 2, participants were given 1,000 generated instances, varying in size. In total, 250 instances with 20, 50, 100 and 200 nodes, respectively. The performance of the proposed algorithms was evaluated on 100 Monte Carlo samples for each instance and averaged over the entire set of instances and samples, i.e., 100,000 simulations. Similarly to Track 1, the performance was measured by the sum of prizes and penalties when evaluating a single Monte Carlo sample and then averaged over all samples and instances. Note that, unlike Track 1, here we are interested in learning a policy that works well for a varying number of instances in the validation set. Participants could use the validation set to check the performance of their policies by utilizing the instances and a specific random seed for comparison to other approaches. 

\paragraph{Test phase}

Only the test phase was used to determine the winners of the competition.
This phase followed a similar procedure as the validation phase. In both tracks, participants had one week to submit their final test scores following the end of the validation phase. In Track 1, a new instance containing $n=65$ nodes was generated to evaluate final performance. Similarly, in Track 2, 1,000 instances were generated in the same fashion as in the validation phase. The procedure for evaluating performance remained unchanged from the validation phase. The instance generating distribution was the same between the validation and test phases in both cases. Thus, an algorithm trained for the validation phase could be used to propose solutions for the instances in the test phase.

\subsubsection{Submissions}

In both tracks, the participants were asked to submit their output files containing the result of their proposed method as well as their implementation code for inspection by the organizing team. In Track 1, the submission file consisted of a single tour. This tour was used to compute the average performance over 10,000 Monte Carlo samples. In Track 2, participants were supposed to submit a single file containing the tours for each instance (1,000) and each Monte Carlo sample (100). In total, 100,000 tours should have been submitted for evaluating performance. This file was then used to compute the overall performance of the submission. In both cases, participants had to submit tours with size $n+1$, where a visit to the depot determined the end of the tour.

\subsubsection{Ranking submissions}

Ranking submissions proceeded as follows. First, we computed the total prize and penalties for each evaluated tour. That is, for a given Monte Carlo sample $j$ and instance $I$, a tour score $ \alpha(\mathbf{s}^{(j)}, I)$ is computed as
\begin{equation}
    \alpha(\mathbf{s}^{(j)}, I) =  \sum_{i=1}^{n} \mathds{1} \Big[s^{(j)}_0 \notin  \{ s^{(j)}_1, \ldots, s^{(j)}_i \} \Big] (p_{s^{(j)}_i} + e_{s^{(j)}_i}),
\end{equation}
where $n = 65$ in Track 1, $n \in \{ 20, 50, 100, 200 \}$ in Track 2, and $\mathds{1}[\cdot]$ is an indicator function that takes the value 1 when the predicate is true and 0 otherwise.
The latter makes sure to stop calculating rewards when the tour returns to the starting depot.
After evaluating the total number of solutions, we average the total performance obtained over all instances and Monte Carlo samples of each track. That is, the final score is given by
\begin{equation}
    \text{score}= \frac{1}{m |\mathcal{I}|}\sum_{I \in \mathcal{I}} \sum_{j =1}^m \alpha(\mathbf{s}^{(j)}, I),
\end{equation}
where $|\mathcal{I}|=1$ and $m=10,\!000$ in Track 1, and $|\mathcal{I}|=1,\!000$ and $m=100$  in Track 2.
Based on this scoring metric, participating teams were ranked in descending order. That is, the teams with higher performance scores were ranked higher.

\section{Winning methods}

This section presents the methods that were used by the winning teams in both tracks of the competition.
Though the organizers had access to the code submitted by the participants, the methods presented in this section are explained by the participants.

\subsection{Track 1 winners}
As there was a three-way tie in Track 1 (see Section~\ref{sec:results}), three methods are presented for this track.
Participants had to make use of surrogate-based optimization techniques to optimize one instance with 65 nodes.

\subsubsection{The Convexers}

The approach we submitted, after significant experimentation, is shaped by three observations: 
\begin{enumerate} 
    \item The expected performance (i.e., the true objective) can be approximated well by using a (sampling based) approximate model for the true probability space and an exact model for an approximation of the true probability space.
    \item A learning surrogate may be used to identify search regions from where a search algorithm can be restarted.
    \item Within a parallel restarted search approach, slight differences in the true objective function approximation may actually help in diversifying the parallel search efforts.
\end{enumerate}

Based on these observations, our approach consists of a parallel portfolio~\cite{isac,cshc} of restarted, hyper-parameterized~\cite{hyper-DS,hyperTS} dialectic search~\cite{dialectic} workers that are tuned using the gender-based genetic algorithm configurator (GGA)~\cite{gga,gga++,tunedGGA}. Each parallel worker uses (its own local selection of) 100 randomly sampled scenarios to approximate the true objective in each local search step. When a worker encounters a solution that may improve its best solution, the variable assignment is evaluated using the exact model for a simplifying approximation of the true probability space. Since all workers use the same approximation, this provides a shared ground for accepting or rejecting a solution as improving among the parallel workers. 

We use a semi-supervised learning approach for forecasting the quality that will be achieved when restarting the search at a new starting point. The unsupervised part of the architecture consists of an auto-encoder that is trained offline to encode and reconstruct permutations into, respectively, from, a compressed latent space. The supervised part is learned online and aims to forecast the quality of the best solution found after restarting from a given permutation. When a worker decides to restart (which is determined by a hyperparameter), a quick local search over the latent space is conducted to find a promising starting point for the next search. This latent vector is decoded, and the resulting permutation is used to restart the search.

This approach solves the test instance in about 90~seconds using roughly 35~minutes of raw compute time over all workers. This efficiency allowed us to also solve all 1,000 instances from track~2 of the competition, which led to a test performance (over the 100 scenarios per instance prescribed in the competition) of 10.8106 (whereby the expected value in the simplified probability space is 10.78, so the scenarios used in the competition evaluation are comparably friendly). Note that this means that using the best default tours (and sticking to them, no matter how the actual transit times evolve) actually outperforms the winning reinforcement solution of Track~2.

\subsubsection{Margaridinhas}

The method proposed by the \emph{Margaridinhas} for Track \#1 of the AI for TSP competition has three main components: 
(1) a mixed-integer surrogate model,
%
(2) an iterative approach, 
%
and (3) a genetic algorithm.
%
The iterative approach improves the surrogate model throughout iterations according to a  fixed set of parameter values, and outputs the best route found at the end.
The genetic algorithm combines solutions output by distinct calls to the iterative approach (with different sets of parameter values) to find even better solutions.


\textbf{(1) Surrogate model.} 
The surrogate model outputs a route that maximizes the deterministic reward plus the estimated penalty based on previous simulations within the iterative approach.
Each node $i \in \mathcal{N}$ in the nodes set $\mathcal{N}$ has a retrievable reward $r_i$,
and each arc $(i,j) \, \forall \ i,j \in \mathcal{N}$ has an estimated penalty $p_{ij}$.
%
%
The decision variable
$x_{ij} \in \{0,1\}$ equals $1$ if arc $(i,j)$ is in the route or $0$ otherwise.
%
%
For the sake of brevity, let $\mathcal{T}$ be the set of values for  $\{x_{ij}\}_{i,j\in \mathcal{N}}$ describing a route
that starts and ends at the depot and does not contain cycles nor subroutes.
Let also \textit{MaxRouteSize} be a parameter controlled externally by the iterative approach that represents the maximum route size.
The surrogate model $\mathcal{M}(\textit{MaxRouteSize})$ is formulated as follows.
\begin{subequations}
    \begin{align}
        \mathcal{M}(\textit{MaxRouteSize}): \quad \max_{x \in \mathcal{T}} \quad &
        \sum_{i \in \mathcal{N}} \sum_{j \in \mathcal{N}} (r_j + p_{ij}) \cdot x_{ij} \label{eq:objective} \\
        \text{subject to:} \quad& \sum_{i \in \mathcal{N}} \sum_{j \in \mathcal{N}} x_{ij} \leq \textit{MaxRouteSize} \label{eq:size}
    \end{align}
\end{subequations}

The objective function \eqref{eq:objective} maximizes the deterministic reward of the route plus the estimated penalty. 
Constraint \eqref{eq:size} bounds the number of arcs in the route to the maximum route size parameter \textit{MaxRouteSize}.
Note that set $\mathcal{T}$ does not enforce the maximum duration $T$ nor the time windows at nodes.
%
%
The surrogate model $\mathcal{M}(\textit{MaxRouteSize})$ learns what arcs to avoid when building a route through better estimated penalties $p_{ij}$ in the objective function, and added cuts according to previously visited solutions.


\textbf{(2) Iterative approach.} 
The iterative approach explores the solution space of the problem according to the parameters, finding distinct solutions accordingly.
The iterative approach has four parameters:
the maximum number of iterations $K$,
the number of simulations per iteration $M$,
the feasibility threshold \textit{FeasibilityThreshold},
and the gap threshold \textit{GapThreshold}.

First, the iterative approach cuts nodes and arcs from the surrogate model $\mathcal{M}(\textit{MaxRouteSize})$ that would never be visited in a feasible route 
(e.g., nodes with a time window starting after the maximum duration $T$), and sets \textit{MaxRouteSize} to $2$.
At each iteration, the approach solves the surrogate model $\mathcal{M}(\textit{MaxRouteSize})$ to obtain a solution $x^k$ and simulates the associated route $s^k$.
If route $s^k$ is feasible in at least \textit{FeasibilityThreshold} percent of the $M$ simulations, the approach stores it and cuts the feasible solution $x^k$ from the feasible region.
Otherwise, the approach cuts the infeasible structure given by the infeasible solution $x^k$ (i.e., a sequence of arcs, excluding the return to the depot, that is most likely never feasible).

If the gap between the best solution found so far and the upper bound for routes with size limited by \textit{MaxRouteSize} is less than \textit{GapThreshold}, the approach increases \textit{MaxRouteSize} by $1$ and starts  the search for larger routes.
The upper bound for routes with size limited by \textit{MaxRouteSize} can be calculated by solving the surrogate model $\mathcal{M}(\textit{MaxRouteSize})$ without penalties $p_{ij}$.
Next, the approach updates the penalties $p_{ij}$ in the objective function according to the penalty of route $s^k$ given by the $M$ simulations.
The total penalty of route $s^k$ is equally divided between its arcs, and the penalty $p_{ij}$ of an arc $(i,j)$ is simply the average over registered penalties.
The algorithm stops and outputs the best route once the best solution meets the upper bound or the approach hits the maximum number of iterations $K$.


\begin{figure}[htb]
\centering
\includegraphics[width=.8\textwidth]{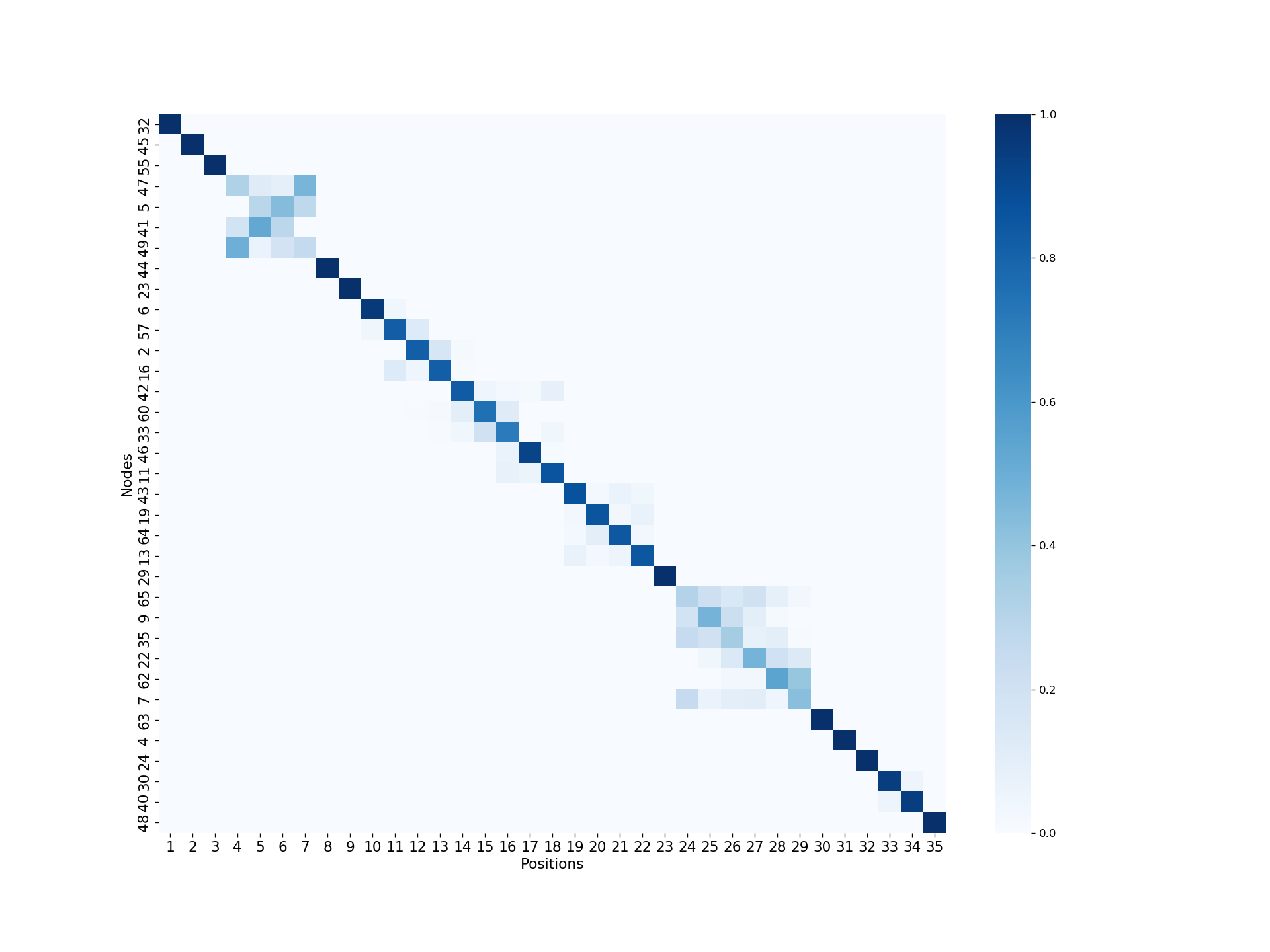}
\caption{Estimation of node probability of occurrence in the positions of optimal solutions. The first visited node is always 32. The nodes visited in 4th to 7th places are \{47, 5, 41, 49\}.}
\label{fig:heatmap}
\end{figure}

\textbf{(3) Genetic algorithm.} A genetic algorithm (GA) is used as a local search policy to improve the set of solutions found through distinct calls to the iterative approach. We run the GA for a few generations on 14 warmed-up solutions output by the iterative approach, with scores ranging from 9.83 to 11.28, to investigate the neighborhood. The iterative approach takes on average 15 minutes to find a warmed-up solution on a Windows computer with an Intel Core i7-10510U CPU @ 1.80GHz × 8 processor and 8 GB RAM memory. The initialization step of the GA uses the 14 warmed-up solutions together with random solutions to make 25600 in total. Even though there were only a few surrogate solutions, they were crucial for the GA to output better solutions rapidly. We ran 5 generations of the GA, performing 100 evaluations for each new solution. After the evaluation of a generation, we reduce the number of solutions to 200 parents by picking the winners of a tournament and the top solutions from past generations. We use the Non-Wrapping Ordered Crossover reproduction operator from \cite{cici2006} since it preserves the order of the nodes. 

The GA has managed to successfully find better solutions than the distinct calls to the iterative approach. We found a number of solutions scoring 11.32, which is the optimal solution for the Track 1 test instance, after running the GA for 2 hours in the previously described setup. \autoref{fig:heatmap} presents the estimated  probability that a node occurs in a certain position of the optimal solution. These estimates are based on the optimal solutions found through the GA. Note that in \autoref{fig:heatmap} the optimal solutions share the exact same set of nodes; moreover, the order of visit of some neighboring nodes can be swapped.
Further details about the method and how it has been implemented can be found on Github
\footnote{https://github.com/almeidawarley/tsp-competition}.

\subsubsection{ZLI}

The approach of the ZLI Team to the competition is based on the assumption that the problem is a black-box and the algorithm only has access to the number of nodes $n$ and the values returned by the objective function $f$. By controlling the number of evaluations per solution used to estimate its expected objective value (Eq.~\ref{eq:objective}), that is, the fidelity of the estimation, it is possible to evaluate hundreds of thousands of solutions within the time limit of the competition. Therefore, from the point of view of classical Bayesian optimization, the problem cannot be considered expensive, and classical approaches based on Gaussian Process regression (GPR), such as CEGO~\citep{ZaeStoFriFisNauBar2014}, are not feasible due to the large number of evaluation points. However, a good estimation still requires thousands of evaluations and, thus, it is important to keep track of each estimated value and its level of fidelity to progressively increase it while avoiding redundant evaluations. Thus, we apply multiple techniques simultaneously to address these limitations:
(1) reducing the dimensionality of the problem by excluding infeasible nodes, (2) caching the expected objective value of evaluated solutions for various fidelity levels, (3) training a classifier to predict the probability of feasibility of a solution, (4) applying a self-adaptive black-box evolutionary algorithm (EA) directly to the objective function while increasing the fidelity, (5) supporting the EA by a surrogate model based on GPR, and (6) enabling the model to deal with many evaluation points via a clustering approach.

As a first step, we reduce the dimensionality of the problem by evaluating all possible tours that visit exactly one node $1\,000$ times. If the result is infeasible in at least 
10\% of the repeated evaluations, the respective node is removed
from the set of available nodes. In the competition instance, this step reduced the dimension from $n=65$ to $n=37$.

Afterwards, we step-wise increase the fidelity of the objective function, 
i.e., increasing the number of repeated evaluations for each candidate tour, in order to find a set of reasonably good solutions. For each of fidelity level of 1, 10 and 100, we run a self-adaptive EA with a budget of $10\,000$ evaluations. The first EA run starts from random tours, and each subsequent run starts from the last population of the preceding (lower-fidelity) run. 

The EA roughly follows the Mixed Integer Evolution Strategy~\cite{Li2013}. Its self-adaption dynamically configures the variation (recombination and mutation) operator choices,
and the mutation rate. The variation operators are mutated randomly with probability $p$, and are recombined by choosing randomly from parents. The mutation rate is itself mutated by: $q^*=q \exp(\tau z)$ with learning rate $\tau$ and $z$ being a sample from the normal distribution (zero mean, unit variance). The mutation rate is recombined via intermediate crossover. The possible choices for variation operators are standard operators for permutations (recombination: Cycle, Order, Position, Alternating Position; mutation: Swap, Insert). We modified these mutation operators to avoid creating solutions that visit less than two nodes (since we already tested all one-node tours during dimensionality reduction) and to avoid mutations that occur in the inactive part of a tour (changing only unvisited nodes).

The EA avoids re-evaluating solutions with the same level of fidelity by maintaining a cache memory of expected values per solution. This cache memory is shared between EA runs. The cache memory after the three initial EA runs is used to train a Gradient Boosting Decision Tree Classifier to predict the probability of a tour being infeasible based on the rank representation of the tour. 

The EA is run a fourth time with a fidelity of 100 and a budget of $10\,000$ using the trained classifier to avoid evaluating solutions if their predicted probability of being feasible is less than $0.4$. 

A fifth run of the EA with a fidelity of 100 and a budget of $5\,000$ incorporates a surrogate model, a Gaussian process regression model (GPR, aka Kriging). At its core, GPR is based on a kernel function that captures the similarity or correlation
between input points, i.e., tours $s$ and $s'$. To enable the GPR model to deal with the discrete sequence points, we use an exponential kernel $k(x,x')=\exp(-\theta d_\text{L}(x,x'))$, where $d_\text{L}(x,x')$ is the Levenshtein distance between two tours. This distance measure counts substitutions, deletions, and insertions of nodes required to turn one tour into another. Only the active part of the tour is considered during distance calculation, i.e., nodes that are actually visited.

Since the surrogate-assisted EA will evaluate thousands of solutions, we need to be able to model large numbers of points. This is usually a challenge for GPR  models, as the required computational effort increases significantly. To deal with this issue, we separate the training data into subsets via clustering, train a GPR model for each cluster, and combine the individual models into an ensemble via a weighted sum.
The ensemble prediction uses the predicted uncertainty of each individual model for weighting. This approach of clustered Kriging is described by Wang et al.~\cite{Wang2017}.
To further save computational effort, the model is only retrained with new data every 20 generations. In each EA generation, the surrogate model is used to pre-filter
the generated offspring, removing the worst 50\%.

We run the EA two additional times. The sixth run simply increases the fidelity to 500. The seventh and final run of the EA uses the probability predicted by the classifier as an additional criterion, together with the prediction of the GPR model, to sort the generated offspring instead of simply discarding solutions that are predicted to be infeasible.

Since each run of the EA starts from the last population of the previous run, the seven EA runs can be seen as different phases of a single overall run of the algorithm, with the shared cache memory acting as a global archive of the best solutions found that is also used to re-train the classifier between runs.

In the last step, the best 250 solutions from the cache memory, which contain the best from the last run of the EA, are evaluated with a fidelity of $10\,000$ evaluations and the best one is returned.

Experiments were carried out on a cluster of 2×8-core Intel Xeon E5-2650v2 2.60GHz with 64GB RAM. We launched multiple independent runs of the above algorithm in parallel to assess the robustness of the results, however, each run is executed sequentially on a single core. A complete run of the above algorithm requires an average of 12 hours to complete.

\subsection{Track 2 winners}

In this track, participants had to use deep reinforcement learning methods to find optimal policies for 1,000 instances with sizes ranging from 20 to 200 nodes.

\subsubsection{RISE up}

Our solution approach for the challenge consists of three components. First, we use the POMO reinforcement learning approach proposed by \citet{pomo} to learn one policy per problem size. Next, we use efficient active search \citep{eas} to fine-tune the learned policies for each instance being solved, thus creating an individualized policy for each instance of the test set. Finally, we use Monte-Carlo rollouts to construct the final solutions. In the following, we will describe each of these three components in more detail.

\paragraph{POMO}
We use POMO  to learn an initial policy for each problem instance size. The POMO approach is an end-to-end deep reinforcement learning approach for combinatorial optimization problems. POMO is based on the REINFORCE algorithm, but exploits symmetries in combinatorial optimization problems to encourage exploration during learning. The network architecture of the employed model is based on the transformer architecture and consists of an encoder and a decoder neural network. The implementation of POMO is made available by the authors. POMO can be used to solve multiple instances in a batch, as well as to perform multiple rollouts per instance in parallel. 

We slightly adjust the POMO implementation to support the TD-OPSWTW. We change the input that POMO expects for each node $i$ to $(x_i, y_i, p_i, l_i, h_i)$. All values are scaled before being input to POMOs deep neural network. We scale the $x_i$ and $y_i$-coordinates based on the 2D-space limits, $p_i$ based on the maximum prize per instance, and $l_i$ and $h_i$ based on the given maximum tour time $T$. Furthermore, we change the decoder context to include the current time $L(\mathbf{s})$ (also scaled by $T$), the embedding of the current node, and the embedding of the depot. Finally, we adjust the masking schema of POMO to forbid actions that correspond to traveling to a node $i$ where $l_i > T$ or where $h_i < L(\mathbf{s})$ as well as previously visited nodes. 

We train one separate policy model for each of the four considered problem sizes. The training set has been generated using the provided instance generator. Each model is trained for several days until full convergence on a single Tesla V100 GPU. For the larger instances with $n=100$ and $n=200$ we use transfer learning and start the training from the policy model trained on $n=50$.

\paragraph{Efficient active search}
Efficient active search (EAS)~\citep{eas} is a method that uses reinforcement learning to fine tune a policy to a single test instance. In contrast to the original active search~\citep{Bello2017NeuralCO}, it only adjusts a small subset of (model) parameters. This allows us to solve multiple instances in parallel, resulting in significantly reduced runtime. For each test instance, we create a copy of the corresponding problem-size specific policy learned by POMO and then fine tune a subset of the policy parameters to that specific instance. This takes up to 30 minutes on our GPU per instance for the larger instances with 200 nodes. The result is a set of 1,000 separate policies. Note that the travel times between nodes are not fixed during the training process. Instead, the travel times are sampled for each solution construction process. This enables EAS to learn a robust policy that can create high-quality solutions for a wide range of scenarios. If we use the fine-tuned policies to construct solutions greedily, we observe an average reward of 10.67 on the test set. In contrast, using the four size-specific models for solution construction results in a reward of 10.43.

\paragraph{Monte-Carlo rollouts}
We construct the solutions for the test instances using Monte-Carlo rollouts and the instance-specific policies learned via EAS. We construct a solution for a test instance as follows. At each decision step, we first use the policy model to generate a probability distribution over all possible actions (i.e., all possible nodes that can be visited next). For the five actions with the highest associated probability values, we then perform 600 Monte-Carlo rollouts each. Each Monte-Carlo rollout starts with  the corresponding actions and then completes the solution by sampling the following actions according to the learned policy model. Once all Monte-Carlo rollouts are finished, the action with the highest average reward is selected. Note that only after the final action has been chosen are the actual travel times between the nodes revealed. For the Monte-Carlo rollouts, travel-times are sampled independently in the preceding step. Using Monte-Carlo rollouts increases the average reward to 10.77 (from 10.67 obtained via a greedy solution using the EAS-based policies).

\subsubsection{Ratel}

 The approach of team \emph{Ratel} relies on a version of the Pointer Networks (PN) designed to tackle problems with dynamic time-dependent constraints, in particular, the Orienteering Problem with Time Windows \cite{GAMA2021}. While the model shares the same basic structure with previous PNs \cite{Bello2017NeuralCO, kool2018attention}, i.e. a set encoding block that encodes each node, a sequence encoding block that encodes the constructed sequence so far and a pointing mechanism block, its architecture differs from previous PNs mainly in that it introduces recurrence in the node encoding step. This recurrence makes it so that both encoding and decoding steps are carried out sequentially for every step of the solution construction process. This aspect brings great advantages, especially when solving problems with dynamic constraints, as it allows the use of masked self-attention using a lookahead induced graph structure, which in turn allows for an updated representation of each admissible node in every step \cite{GAMA2021}.

\paragraph{Input Features} 
Since the model is recursive, at each step it determines a feature vector associated with every admissible node. These feature vectors are given as input to the set encoder in order to compute a step-dependent representation of each node. This feature vector is a combination of static features -- that remain constant throughout the solution construction process -- and dynamic features -- that can change at every step. For this problem, the model uses $11$ static features and $34$ dynamic features.

The static features are obtained directly from each instance data. Concretely, the Euclidean coordinates of each node $i$ ($x_i$ and $y_i$), opening and closing time ($l_i$ and $h_i$), time window width ($h_i - l_i$), prize ($p_i$), maximum time available (i.e. max length, $T$), prize divided by time window width, prize divided by distance from depot, and difference between maximum time available and opening and closing times ($T-l_i$ and $T-h_i$). As for dynamic features, the model uses $34$ features that are functions of the current time and current node. Some of the dynamic features are boolean and indicate whether a node's feasibility conditions are satisfied, assuming either the fastest travel times possible or the worst travel times. Some examples of non-boolean dynamic features are, for each node, the time left until the opening time, the time left until closing time, the fraction of time elapsed since tour start, prize divided by max time to arrive to the node, prize divided by time to closing time, the probability of arriving after closing time, the probability of arriving after the maximum time available and the expected prize of the node.

\paragraph{Setup and Training} 

In order to speed-up training and model development, training was done by sampling from 804000 pre-generated instances: 4000 instances for each number of nodes between 10 and 210 nodes, with seeds ranging from 1 to 4000. During training, at each step, the model samples one instance from the pre-generated set of instances (one number of nodes between $10$ and $210$ and one seed between $1$ and $4,\!000$) with replacement. Note that the seeds from the validation phase ($12345$) and the test phase ($19120623$) are outside this range and thus were not considered during training. This is to avoid/minimize leakage and overfitting and achieve a more realistic final collected prizes.

The model was trained for $15,\!000$ epochs, using a method based on the REINFORCE algorithm \cite{Williams1992} as in \cite{GAMA2021}, with entropy regularization to enhance exploration, $L_2$ regularization and Adam optimizer. In each epoch, $6$ simulations on the same instance were performed, each one composed of $32$ sample solutions, equaling a batch size of $192$ sampled solutions. Further implementation details and hyperparameter values can be found in GitHub \footnote{\url{https://github.com/mustelideos/td-opswtw-competition-rl}}.

All experiments were performed on a 6 core CPU at 1.7 GHz (Intel Xeon CPU E5-2603 v4) with 12 GB of RAM and an Nvidia GeForce GTX 1080 Ti GPU. The final model took up to $48$ hours to train and around $12$ hours to generate the validation/test submission files. 

\section{Results and discussion}\label{sec:results}

This section presents the results of all participating teams in the validation phase and final test phase.
See Table~\ref{tab:results} for the score at the deadline of each phase. Only the test phase was relevant for deciding on the winners of the competition.
As can be seen, nine teams participated in the validation phase and seven in the test phase.
The leaderboard with results was visible to all participants and was updated every time a participating team submitted a solution.
During the validation phase, we saw a large improvement in the scores, indicating that teams were actively improving their methods.
It can also be seen that at the end of the test phase, there was a three-way tie for the top participants of Track 1.
Although this was not the case from the beginning of this phase, during the test phase, these three participants managed to find the globally optimal solution of the instance under consideration.
The prize money for first and second place of Track 1 was split among these three teams (ZLI, Margaridinhas, and Convexers), while the prize money for first and second place of Track 2 was given to teams RISEup and Ratel, respectively.

\begin{table}[tbp]
    \caption{Leaderboard for the validation phase and the final test phase at the deadline.}
    \label{tab:results}
    \centering
    \begin{tabular}{cc|cc}
    \toprule
        Track 1 (SBO) & Validation phase & Track 1 (SBO) & Test phase\\
        \midrule
         ZLI & 8.404499999999793 &  Convexers & 11.320000000002786
         \\
         Convexers & 7.8099999999988885 & Margaridinhas & 11.320000000002786\\
         Margaridinhas & 7.533834999999982 & ZLI & 11.320000000002786  \\
         Topline & 3.540000000000663 & Topline & 4.300000000000301\\
         VK & 0.5300000000000102 & - & -\\
         \midrule
         Track 2 (DRL) & Validation phase & Track 2 (DRL) & Test phase\\
         \midrule
         Ratel & 10.69104 & RISEup & 10.77341\\
         ML for TSP & 9.82955 & Ratel & 10.58859\\
         RISEup & 9.16567 & ML for TSP & 10.39341\\
         VK & -6.69772 & - & - \\
         UniBw & -13.14861 & - & - \\
    \bottomrule
    \end{tabular}
    
\end{table}












We have obtained several insights from organizing this competition. First of all, our main goal of advancing the state-of-the-art of ML methods in routing problems has been achieved, with seven new surrogate or reinforcement learning methods presented in this work by the winning participants, and other methods having been applied by the other participants.
Our other goals, namely attracting ML researchers to solve a challenging routing problem, and creating a new routing problem as a benchmark for ML, have also been achieved, making this a successful competition in the eyes of the organizers.

What we noticed was that having clear restrictions on what is and what is not allowed in the competition stimulated creativity.
By limiting participants to use SBO and DRL approaches, new interesting methods have been developed.
Another aspect of the competition that was beneficial for everyone was a chat function, where any confusion was quickly cleared up.

There are also points of improvement. It remains questionable whether providing the whole simulator to participants is really the way to go in an ML competition. Although the samples of random variables in the simulator were not known in advance, the exact probability distributions could be retrieved by looking at the code to which participants had access. This was known by the organizers, but it was decided to not obfuscate this information so that it remained easy for participants to run the provided simulator and use it for ML.

Another point of improvement is the difficulty of the problem, especially for Track 1. Because we did not want to slow down the solution evaluation process on our server too much, and because we noticed solutions in the validation phase were still being improved, we decided to keep the instance size somewhat limited.
Over the course of the test phase, the methods of the participants were being improved up to the point that three of the methods managed to find the globally optimal solution, causing a three-way tie. Although this shows the power of these three different methods, for the purpose of the competition, it is best if the organizers do not underestimate the participants and instead are inclined towards too difficult instances. This should hopefully push the creativity of participants and the power of their methods even further.

For future work, we will consider the points of improvement and see whether the winning methods can be generalized to be used in other routing problems.

\section{Conclusion}
We reported on the first international competition on AI4TSP. The participants were asked to solve a time-dependent orienteering problem with stochastic weights and time windows (TD-OPSWTW). We described the setup of two tracks, focusing on two learning techniques from AI: surrogate-based optimization and deep reinforcement learning. We described the approaches of the winning teams of two tracks and gave an overview of the results. Furthermore, we explained the developed simulation model, which was used to generate the training and testing instances of TD-OPSWTW, and was coupled with two implemented baseline algorithms of two tracks. In addition, the simulator used in this competition and the code of some of the winning approaches of both tracks of the competition has been made publicly available. The simulation model with various algorithms can serve as a benchmark for researchers to develop and compare surrogate-based and reinforcement learning-based approaches to stochastic routing problems. 

This paper was written together by the organizers and the winning teams of the competition. The organizers were pleased with the outcome, as the purposes of the first AI4TSP competition have been achieved. The results show the diversity of the adopted methods, from pure machine learning approaches to integrating learning with more traditional heuristics. There is great potential for developing other sophisticated algorithms, especially on leveraging machine learning into expert-crafted heuristics, to solve (stochastic) routing problems. 

Looking forward, the organization team will continue AI4TSP by considering various improvements such as setting time and computation budget and implementing more realistic distribution functions in the simulator. In addition, other practical and societal relevant optimization problems will be considered in the future editions of the competition.

\section{Acknowledgement}
The organizers would like to thank Ortec and Vanderlande for sponsoring the prize money. 
\bibliography{mybibfile}

\end{document}